\documentclass[runningheads]{llncs}
\usepackage{graphicx}
\usepackage{subfigure}
\usepackage{cite}
\usepackage{amsmath,amssymb}
\usepackage{booktabs}
\usepackage{multirow}
\usepackage{amssymb}
\usepackage{booktabs}
\usepackage{tabu}
\usepackage{caption}
\usepackage{subfigure}
\usepackage{amsmath}

\begin{document}
\bibliographystyle{unsrt}
\title{AFO-TAD: Anchor-free One-Stage Detector for Temporal Action Detection}

\author{Yiping Tang \and
Chuang Niu \and
Minghao Dong \and
Shenghan Ren \and
Jimin Liang*}

\authorrunning{Y. Tang et al.}

\institute{Engineering Research Center of Molecular and Neuro Imaging of Ministry of Education, School of Life Science and Technology, Xidian University, China\\
\email{jimleung@mail.xidian.edu.cn}}

\maketitle

\begin{abstract}
Temporal action detection is a fundamental yet challenging task in video understanding. Many of the state-of-the-art methods predict the boundaries of action instances based on predetermined anchors akin to the two-dimensional object detection detectors. However, it is hard to detect all the action instances with predetermined temporal scales because the durations of instances in untrimmed videos can vary from few seconds to several minutes. In this paper, we propose a novel action detection architecture named anchor-free one-stage temporal action detector (AFO-TAD). AFO-TAD achieves better performance for detecting action instances with arbitrary lengths and high temporal resolution, which can be attributed to two aspects. First, we design a receptive field adaption module which dynamically adjusts the receptive field for precise action detection. Second, AFO-TAD directly predicts the categories and boundaries at every temporal locations without predetermined anchors. Extensive experiments show that AFO-TAD improves the state-of-the-art performance on THUMOS'14.
\end{abstract}

\section{Introduction}
In recent years, deep convolutional network has become a powerful tool in action recognition, which aims to classify the categories of manually trimmed video clips \cite{simonyan2014two,tran2015learning,hara2017learning,qiu2017learning,feichtenhofer2018slowfast}. However, long untrimmed videos are much more commonly available than trimmed videos in the wild. This phenomenon motivates the study of temporal action detection task which requires not only recognizing, but also localizing the boundaries of action instances in untrimmed video streams.

Many state-of-the-art methods for temporal action detection are akin to object detection architectures \cite{xu2017r,lin2017single,chao2018rethinking}, due to the similarity between the two tasks. Generally, these methods can be divided into two-stage and one-stage frameworks. The two-stage framework, also called ``detection by classification'', aims to generate temporal candidate proposals first, and classify them with separate video classifiers. Although the two-stage architectures have achieved state-of-the-art performance on several benchmarks, they usually suffer from unsatisfying inference speed. Moreover, the strategy that trains the proposal and classification stage separately, prevents collaboration between the two modules \cite{zhang2018s3d,buch2017end}. One-stage architectures detect action instances in untrimmed videos without proposal generation, which makes them operate efficiently. Both of the architectures tend to use predetermined anchors for prediction. However, the performance of anchor-based method is sensitive to the design of anchors. Even with careful design, the detector with fixed temporal scales encounters difficulties in detecting action instances with large variations in duration.

\begin{figure}[t]
\centering
\includegraphics[width=0.7\columnwidth]{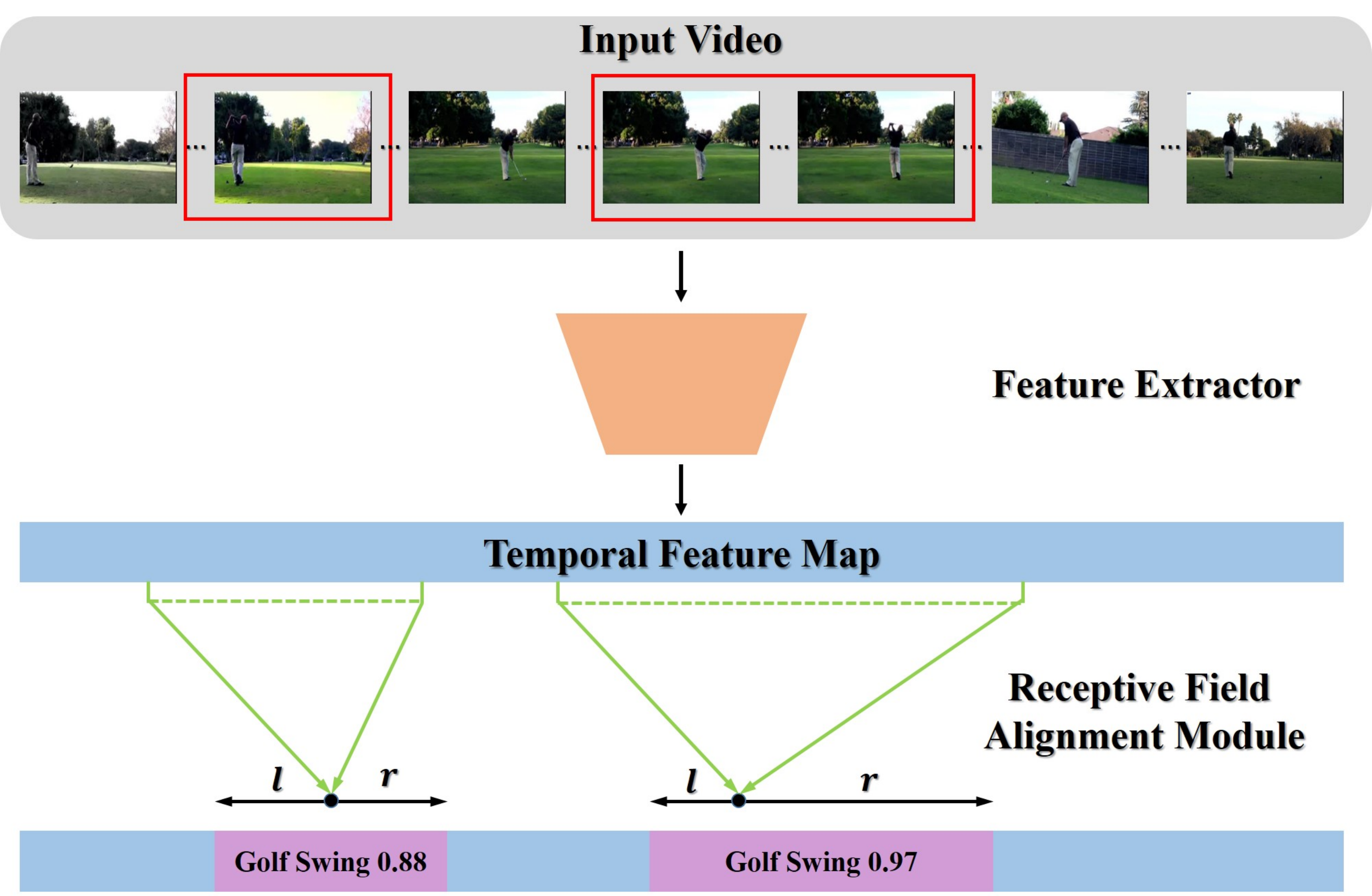}
\caption{The inference process of AFO-TAD. The temporal feature map of input video is extracted by a 3D network. The receptive field adaption module predicts a vector $(c,l,r)$ for each temporal location of video feature, with different size of receptive field. $l$ and $r$ indicate the distances between the starting and ending time to the current temporal location respectively, and $c$ denotes the classification score of the instance.}
\label{fig1}
\end{figure}

To address these issues caused by anchor mechanism, anchor-free detectors such as CornerNet \cite{law2018cornernet}, fully convolutional one-stage object detector (FCOS) \cite{tian2019fcos} and others \cite{zhu2019feature,kong2019foveabox}, have been proposed and superior performances have been achieved in several object detection benchmarks. Due to the fact that the durations of action instances vary from few seconds to several minutes, we argue that the predetermined anchors, which produce fixed receptive field while predicting, is not the optimal choice for temporal action detection. In order to detect action instances with arbitrary durations, we propose a novel architecture named anchor-free one-stage temporal action detector (AFO-TAD). AFO-TAD predicts the categories and boundaries of action instances in untrimmed videos simultaneously as shown in Figure \ref{fig1}. Given a long untrimmed video, the first step is to generate input video clips by sliding windows, whose lengths are only limited by the GPU memory. Then a 3D network (e.g., C3D \cite{hara2017learning}, P3D \cite{qiu2017learning}) is utilized as the feature extractor to generate the robust video feature map. The video feature map is fed into the receptive field adaption module (RFAM) to predict the categories and boundaries at every temporal locations without predetermined anchors. AFO-TAD makes its predictions with suitable contextual information by dynamically adjusting receptive field in RFAM, which is beneficial to action detection performance. The whole network is trained in an end-to-end manner by optimizing the joint loss (classification loss and localization loss).

The contributions of this paper are summarized as follows:
\begin{itemize}
\item We introduce AFO-TAD, an anchor-free one-stage temporal action detector, to predict the categories and boundaries of action instances in untrimmed videos simultaneously.
\item We investigate the effect of receptive field to the performance of one-stage action detectors and find that the dynamic receptive field significantly improves the performance.
\item AFO-TAD achieves state-of-the-art performance on THUMOS'14 benchmark \cite{jiang2014thumos}.
\end{itemize}

\section{Related work}
\subsection{Temporal Action Detection.} Temporal action detection requires both of the categories and boundaries of action instances in untrimmed videos. Generally speaking, existing works can be divided into two-stage and one-stage categories. The two-stage methods firstly generate proposals which are likely to contain the instances of interest, and then recognize them with separate classifier. With the advent of deep learning, the video classifiers for trimmed videos have achieved great success \cite{simonyan2014two,tran2015learning}. Due to the advanced performance of video classification, many researchers regard the proposal generation algorithm as the bottleneck of temporal action detection. However, the two-stage methods often generate proposals by using sliding windows \cite{karaman2014fast,oneata2014lear,wang2014action} with various sizes and ratios, which suffers from low localization precision and high computational burden. Although some improved proposal generation methods have been presented, such as \cite{escorcia2016daps,caba2016fast,lin2018bsn}, the strategy of treating proposal generation and classification as two separate and sequential processing stages may lead to repeated computation and sub-optimal performance.

To address these issues of two-stage methods, one-stage action detectors like SSAD \cite{lin2017single}, SS-TAD \cite{buch2017end} and S3D \cite{zhang2018s3d} have been proposed. Inspired by SSD \cite{liu2016ssd}, SSAD directly localizes the instances by temporal convolution with predetermined anchors. Although the anchor-based one-stage architectures train the classification and localization jointly, they always suffer from inferior performance compared with two-stage architectures. We argue that this is due to the fact that the video classifiers of two-stage methods only operate on the action proposals, thus the receptive fields of video classifiers are aligned to the action instances. While the one-stage methods produce their detection results with fixed temporal scales. Most recently, GTAN \cite{long2019gaussian}, a one-stage action detector which achieves the state-of-the-art performance, learns the temporal structure through Gaussian kernels without predetermined anchors. In this work, we demonstrate that one-stage methods can gain a large margin improvement by eliminating fixed temporal scales.

\subsection{Object Detection.} Object detection and temporal action detection are closely related, for the former detects object by predicting spatial bounding boxes, while the latter detects actions in temporal domain. Faster R-CNN \cite{ren2015faster} proposes a region proposal network (RPN) to generate object proposals and classify them in an end-to-end way. SSD \cite{liu2016ssd} directly outputs the bounding boxes and classification scores in a single pass way without the proposal generation stage. Recently, anchor-free methods \cite{law2018cornernet,zhu2019feature,kong2019foveabox}, which aims to address the issues caused by predetermined anchors, have shown superior performance for object detection. FCOS \cite{tian2019fcos} is the typical architecture of anchor-free object detectors, which eliminates the anchors by predicting the distances between the center point to the four edges of the bounding box. In order to suppress low-quality bounding boxes, FCOS proposes a center-ness mechanism to down-weight the predictions far away from the center of objects.

Inspired by these object detection methods, many powerful action detection architectures have been proposed \cite{xu2017r,lin2017single,chao2018rethinking,zhang2018s3d}.

\begin{figure*}[t]
\centering
\includegraphics[width=0.92\textwidth]{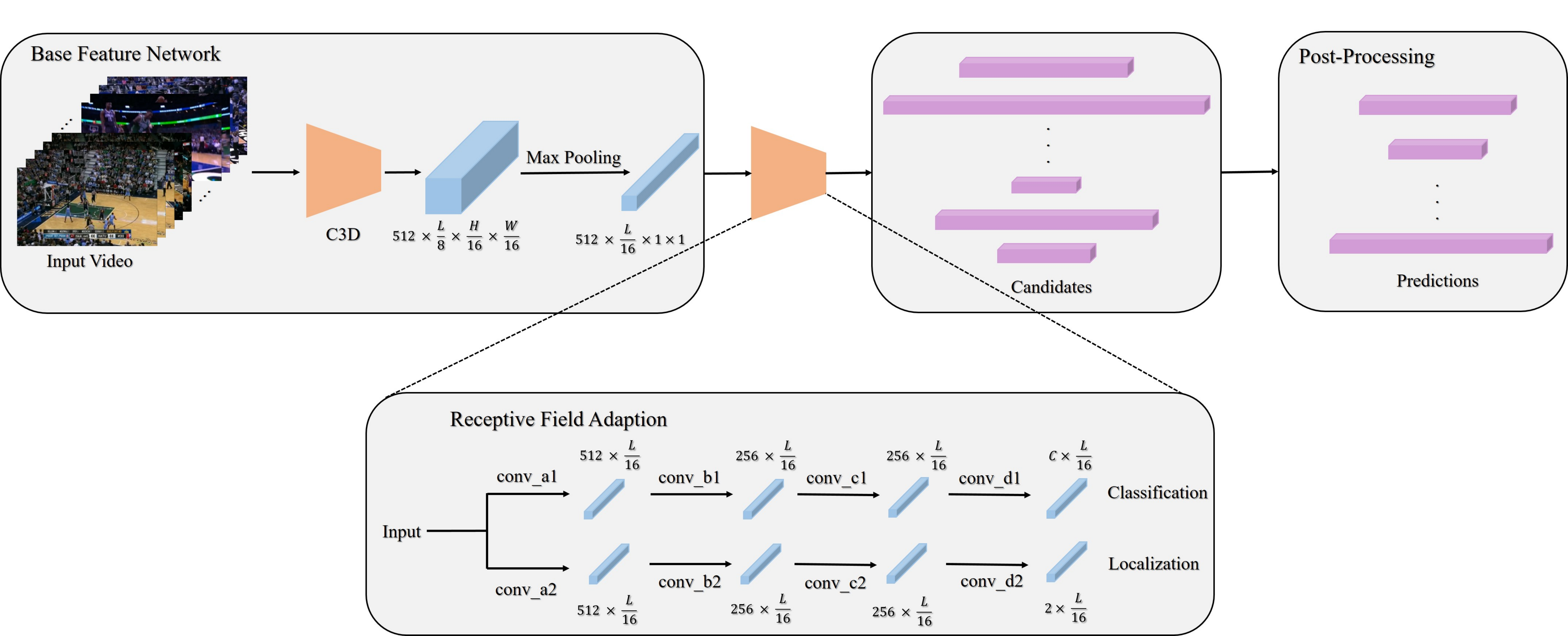} 
\caption{Overview of our anchor-free one-stage temporal action detector (AFO-TAD) architecture. The input of AFO-TAD is a set of RGB frames with the  dimension of $3 \times L \times H \times W$, where $L$ is the length of video clip, $H$ and $W$ are the height and width of each frame. The video feature map is generated by a C3D network followed by a max-pooling layer. The feature map is fed into the receptive field adaption module (RFAM) which consists of two separate branches. The classification branch generates the classification scores $c' \in \mathbb{R}^C$ for each temporal location, where $C$ is the number of categories plus one background and the localization branch outputs two location offsets ($l',r'$) to regress the starting and ending time of the action instance. The action candidates are generated by transforming the location offsets obtained from RFAM. Finally the action predictions are produced by filtering the classification scores and non-maximum suppression (NMS) operation.}
\label{fig2}
\end{figure*}

\subsection{Receptive Field.} Many works have demonstrated that temporal action detection task will benefit from additional contextual information \cite{zhao2017temporal,chao2018rethinking}. In order to get larger receptive field, two-stage methods can simply extend the beginning and ending of proposal regions \cite{chao2018rethinking}, while one-stage methods need to design the receptive field carefully. Down-sampler like pooling layers and dilated convolution are the common ways to enlarge the effective receptive fields (ERF) for one-stage methods. However, the ERF only takes a small portion of the theoretical receptive field \cite{luo2016understanding}, which suggests that it's hard for one-stage methods to select the suitable receptive field. The simplest way to alleviate the ERF problem is by using the feature pyramid architecture \cite{lin2017feature}, which predicts action instances with various durations on layers with different receptive field. Nevertheless, the feature pyramid architecture hinders the application of network input with high temporal resolution due to the fact that the computation cost will increase rapidly with the number of input frames (e.g., in \cite{zhang2018s3d}, the input video clips are decoded at 8 frames per second).

In order to get the dynamic receptive field, the deformable convolution \cite{dai2017deformable,zhu2019deformable} for object detection learns the spatial sampling locations with additional offsets. This operation changes its ERF by dynamically adjusting the sampling positions during training. In this paper, we apply temporal deformable convolution for action detection and demonstrate that the anchor-free one-stage architecture benefits significantly from the dynamic receptive field. As a result, our method can operate on high temporal resolution video clips decoded at 25 frames per second.

\section{Our Approach}
In this section, we will introduce the anchor-free one-stage temporal action detector (AFO-TAD), a novel architecture for temporal action detection in long untrimmed videos. The architecture, as illustrated in Figure \ref{fig2}, consists of two components: a base feature network and a receptive field adaption module (RFAM). Given a video clip $V=\{f_t\}_{t=1}^{L}$ ($L$ is only limited by the GPU memory), the temporal feature map is generated by a 3D convolutional network followed by a max-pooling layer. Then the temporal feature is fed into RFAM, which consists of two branches for action classification and boundary localization, respectively. The RFAM generates the classification scores and location offsets at every temporal positions in the feature map. In order to get the dynamic receptive field, $conv\_d1$ and $conv\_d2$ in RFAM are implemented by using the temporal deformable convolution. The whole network is optimized by a joint loss of classification and localization in an end-to-end manner.

\subsection{Base Feature Network.} We use C3D \cite{hara2017learning} to extract the robust feature map of input video clip. The input to our network is a sequence of RGB frames with the dimension of $3 \times L \times H \times W$, where $L$ is the number of frames, $H$ and $W$ are the height and width of each frame. The feature extractor is composed by the $conv1a$ to $conv5b$ layers of C3D and a 3D max-pooling layer (kernel size $2 \times \frac{H}{16} \times \frac{W}{16}$, temporal stride 2). The 3D max-pooling layer is applied to produce a temporal feature map with the dimension of $512 \times \frac{L}{16} \times 1 \times 1$. For simplicity, the spatial dimension of 1 $\times$ 1 is omitted in Figure \ref{fig2}. More experiments will be shown in next section to indicate the necessity of the down-sampler (i.e., the max-pooling layer).

\subsection{Receptive Field Adaption Module.} 
\begin{figure}[t]
\centering
\subfigure[Temporal dilated convolution]{
\begin{minipage}{0.7\columnwidth}
\centering
\includegraphics[width=1\columnwidth]{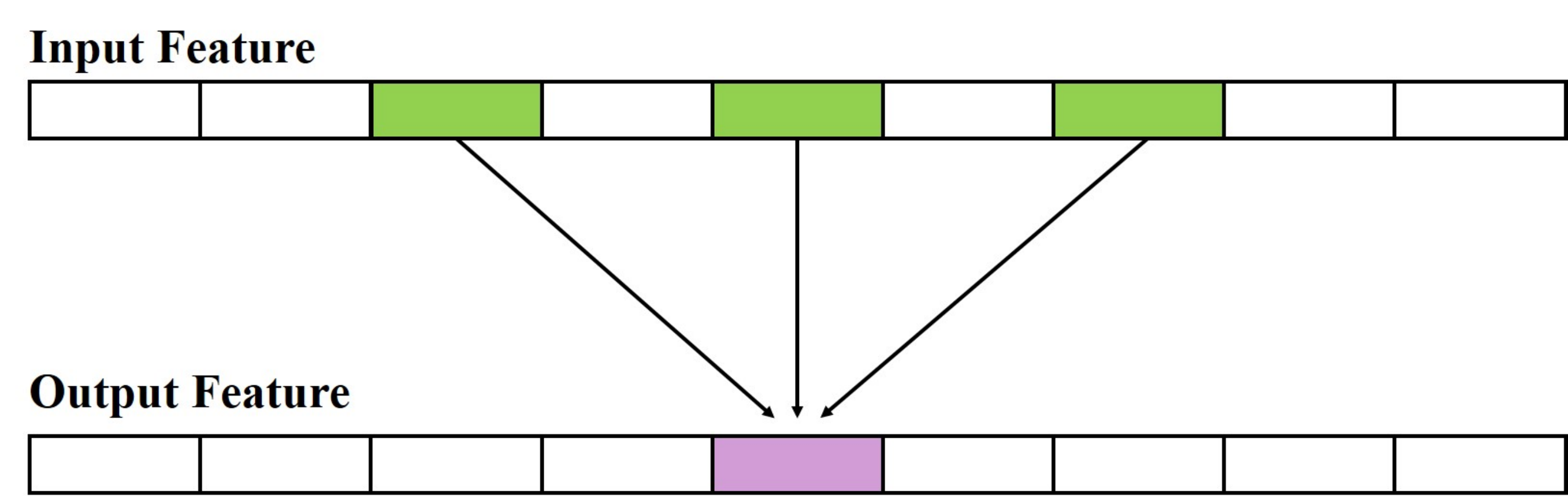}
\vspace{-2mm}
\end{minipage}
}
\vspace{1mm}\\[4mm]
\subfigure[Temporal deformable convolution]{
\begin{minipage}{0.7\columnwidth}
\centering
\includegraphics[width=1\columnwidth]{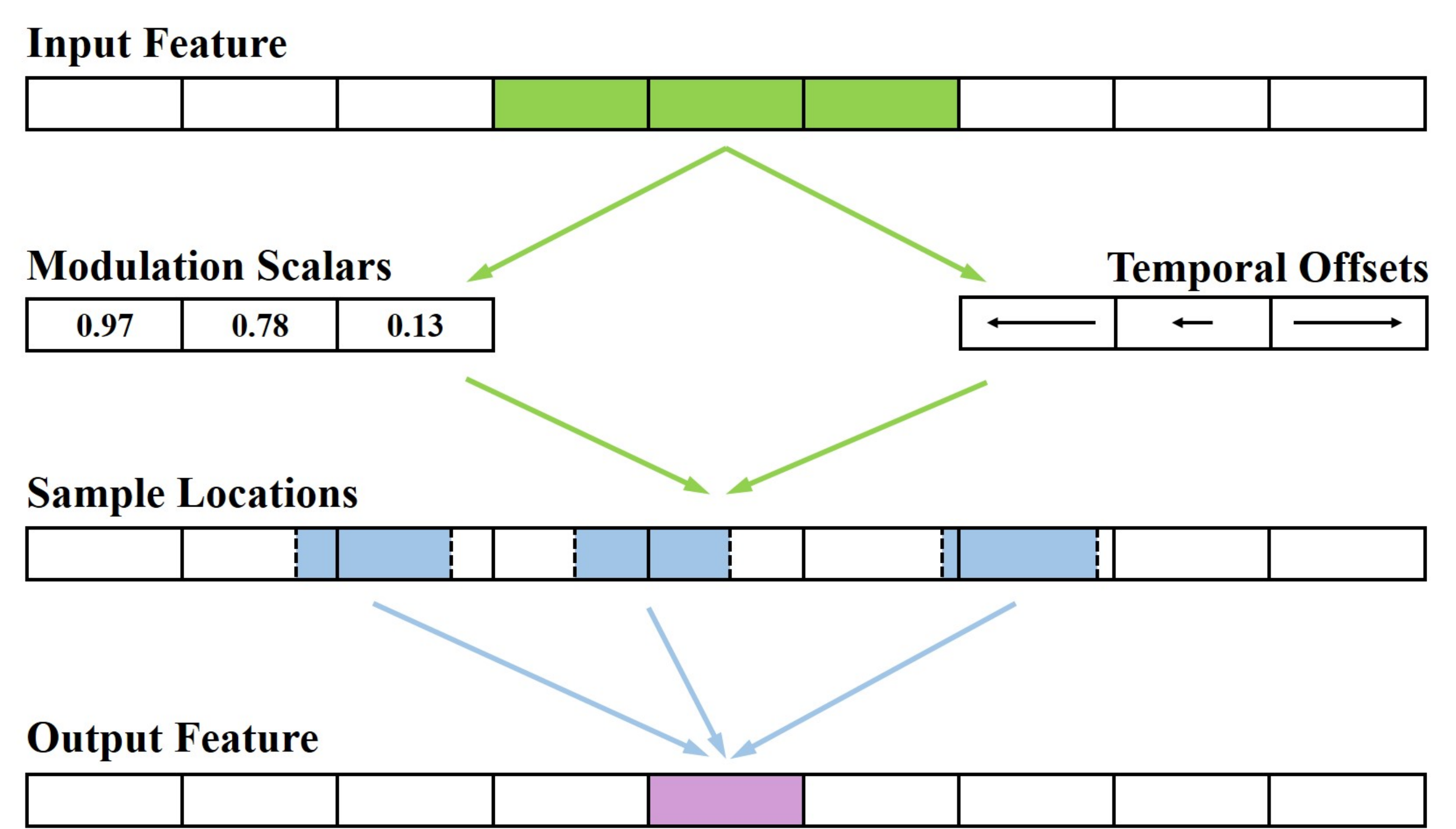}
\vspace{-2mm}
\end{minipage}
}
\caption{Illustration of (a) temporal dilated convolution and (b) temporal deformable convolution with the kernel size of 3. The sample locations of temporal dilated convolution is pre-specified by the dilation rate (2 in the figure), which leads to a fixed receptive field. Whereas the sample locations of temporal deformable convolution is further adjusted by the learnable offsets. Together with the learnable modulation scalars, temporal deformable convolution adjusts its effective receptive field adaptively.}
\label{fig3}
\end{figure}

The purpose of RFAM is to generate the classification scores and location offsets for each temporal feature point with dynamic receptive field. The output of base feature network is $F\in\mathbb{R}^{512 \times \frac{L}{16} \times 1 \times 1}$. The $conv\_d1$ and $conv\_d2$ in RFAM are temporal deformable convolutional layers while the others are normal temporal convolutional layers. The temporal deformable convolution is illustrated in Figure \ref{fig3}, which can be formulated as \cite{zhu2019deformable}:
\begin{equation}\label{eq0}
  y(p)=\sum_{k=1}^{K}w_k \cdot x(p+p_k+ \Delta p_k) \cdot \Delta m_k,
\end{equation}
where $x(p)$ and $y(p)$ are the values of input and output feature maps at temporal location $p$ respectively. $K$ is the kernel size and $w_k$ the kernel weights. $p_k$ is the pre-specified offset for the $k$-th kernel position. $\Delta p_k$ and $\Delta m_k$ are the learnable offset and modulation scalar respectively. The sampling location of the deformable convolution is $p+p_k+ \Delta p_k$ while that of the normal convolutional layer is $p+p_k$. The modulation scalar is a weighting term used to control the contribution of each temporal location. For action instances with different durations, the temporal deformable convolutional layer is able to adjust its receptive field dynamically, which is benefit to the detection performance.

RFAM generates the classification scores and location offsets for each temporal feature location $x$. The outputs of RFAM can be denoted as $\{o_x=(c',l',r')\}_{x=1}^\frac{L}{16}$. $c'\in\mathbb{R}^{C}$ is the class score vector and $C$ the number of categories plus one (background class). $l'$ and $r'$ are the location offsets.

\subsection{Prediction and Post-processing.} For each temporal feature location $x$, an action candidate is generated based on the outputs of RFAM. The classification score $c\in\mathbb{R}$ for the candidate is calculated by a $softmax$ function applied to $c'$. The starting and ending time offsets of the candidate are defined as:
\begin{equation}\label{eq1}
  l=exp(\alpha\cdot l'),
\end{equation}
\begin{equation}\label{eq2}
  r=exp(\alpha\cdot r'),
\end{equation}
where $\alpha$ is a trainable parameter. Thereafter, the starting time $s$ and ending time $e$ of the action candidate are obtained as:
\begin{equation}\label{eqs}
  s=x-l,
\end{equation}
\begin{equation}\label{eqe}
  e=x+r.
\end{equation}

The final action predictions are generated by firstly filtering the candidates (thresholding the classification scores with a predefined threshold) and then applying the non-maximum suppression operation (NMS) on the remained candidates.

\subsection{Training.} Given a long untrimmed video, the input video clips are generated by sliding window with the length of $L$ and the overlap ratio of 25\%. The common sliding window method scans the video from the beginning to the end. In order to make the model more robust and get more training data, we employ the ``two-way buffer'' method proposed in \cite{xu2017r}. The two-way buffer method adds another way by sliding window from the end of the video to the beginning. Then the video clips containing (completely or partially) the action instances are selected for training.

The ground-truth instances are denoted as $\{g_i=(\hat{c},\hat{s},\hat{e})\}_{i=1}^m$ where $m$ is the total number of instances in the video clip, $\hat{c}$, $\hat{s}$ and $\hat{e}$ denote the category, starting time and ending time for the $i$-th instance, respectively. We assign instance $g_i$ to temporal location $x$ only if $x\in(\hat{s}, \hat{e})$. However, $x$ may fall into multiple ground-truth instances which makes the location $x$ an ambiguous sample. Fortunately, unlike the object detection problem, most action instances are non-overlapping in temporal action detection. The ambiguous samples only occupy a small portion of the total training data (about 5\% in THUMOS'14). Therefore, we simply assign location $x$ to the instance whose center point is most close. More formally, if location $x$ is associated to $g_i$, it is defined as a positive sample. Its classification target is $\hat{c}$ and the regression targets are formulated as:
\begin{equation}\label{eq3}
  \hat{l}=x-\hat{s},
\end{equation}
\begin{equation}\label{eq4}
  \hat{r}=\hat{e}-x.
\end{equation}

The overall loss function is a weighted sum of classification and localization losses:
\begin{equation}\label{eq5}
  loss=loss_{cls}+\beta\cdot loss_{loc},
\end{equation}
where $\beta$ is a weight term used for balancing each part of the loss function. $loss_{cls}$ is a standard $softmax$ loss over multiple classes confidences ($c'$). $loss_{loc}$ is a temporal IoU loss which is similar to the IoU loss used in UnitBox \cite{yu2016unitbox}. The temporal IoU loss is shown as:
\begin{equation}\label{eq6}
   loss_{loc}=-\frac{1}{N_{pos}}\sum_{i=1}^{N_{pos}}ln(\frac{I_i}{U_i}),
\end{equation}
where
\begin{equation}\label{eq7}
   \begin{split}
      I_i & =min(\hat{l},l)+min(\hat{r},r), \\
      U_i & =(l+r)+(\hat{l}+\hat{r})-I_i,
   \end{split}
\end{equation}
and $N_{pos}$ is the number of positive samples. The whole network is trained in an end-to-end manner by penalizing the two losses.

\subsection{Inference.} Given a long untrimmed video, the video clips are generated by sliding window from the beginning of the video to the end with the overlap ratio of 25\%. Different from the training stage, all of the video clips generated by sliding window are retained in the inference stage. The following inference steps have shown as aforementioned.

\section{Experiments}
We evaluate AFO-TAD on THUMOS’14 \cite{jiang2014thumos}, a large-scale action detection benchmark. The experimental results show that our network achieves the state-of-the-art performance.

\subsection{Dataset.} THUMOS'14 is a widely used dataset which contains over 20 hours of video from 20 action categories for temporal action detection task. The validation set and test set contain 200 and 213 untrimmed videos respectively. Following the standard practice, the validation set is used for training and the test set is used for evaluation. Results are reported following the official metrics used in THUMOS'14. The Average Precision (AP) is computed for each action category and the mean Average Precision (mAP) with different temporal IoU thresholds is calculated.

\subsection{Experimental Setup.} The videos in our work are decoded at 25 frames per second (FPS). Although the length of input can be arbitrary, AFO-TAD takes $L=768$ frames as input with the spatial size of 112$\times$112 due to the limited GPU memory. The choice of input length is based on the fact that over 99.5\% action instances in the training set have smaller length than 30.7 seconds. The video clips are generated by sliding window with the overlap ratio of 25\%.  In order to demonstrate the advantages of the proposed anchor-free one-stage action detection method, C3D is chosen as our base feature network instead of more complex models such as P3D. We initialize the base feature network with C3D weights pre-trained on Sports-1M by the authors \cite{hara2017learning} and other layers from scratch. Considering the training efficiency, we fix the weights of the first two convolutional layers of base feature network. The initial learning rate is set as 0.0006, and decreased by 10\% after every 20k iterations. The mini-batch size is 4 and the weight decay parameter is 0.0005. The balance parameter $\beta$ in Equation (\ref{eq5}) is set to 1. After AFO-TAD generates action candidates for a video clip, the candidates with classification scores smaller than 0.005 are filtered. Then NMS with threshold 0.3 is applied and at most 300 top-scoring candidates are selected as the final action predictions. Our networks are trained by utilizing stochastic gradient descent (SGD) with momentum of 0.9.

\begin{table*}[!t]
  \huge
  \centering
  \scriptsize
  \caption{Action detection results (mAP) on THUMOS'14 with various IoU threshold $\theta$. The results show that the proposed AFO-TAD achieves the state-of-the-art performance.}
  \label{tb1}
  \begin{tabu}to 0.7\textwidth{X[0.1mm,c] *6{X[-1, c]}}
    \\[-2mm]
    \toprule\\[-2.5mm]
    & $\theta$ & 0.3 & 0.4 & 0.5 & 0.6 & 0.7\\[0.5mm]
    \hline
    \vspace{1mm}\\[-2mm]
    \multirow{7}*{Two-stage Methods} & SCNN \cite{shou2016temporal} & 36.3 & 28.7 & 19.0 & 10.3 & 5.3\\
    \vspace{1mm}\\[-2mm]
    & TURN \cite{gao2017turn} & 44.1 & 34.9 & 25.6 & 14.6 & 7.7\\
    \vspace{1mm}\\[-2mm]
    & CDC \cite{shou2017cdc} & 40.1 & 29.4 & 23.3 & 13.1 & 7.9\\
    \vspace{1mm}\\[-2mm]
    & R-C3D \cite{xu2017r} & 44.8 & 35.6 & 28.9 & - & -\\
    \vspace{1mm}\\[-2mm]
    & CTAP \cite{gao2018ctap} & - & - & 29.9 & - & -\\
    \vspace{1mm}\\[-2mm]
    & BSN \cite{lin2018bsn} & 53.5 & 45.0 & 36.9 & 28.4 & 20.0\\
    \vspace{1mm}\\[-2mm]
    & BMN\cite{lin2019bmn} & 56.0 & 47.4 & 38.8 & 29.7 & \bf{20.5}\\[-2mm]\\
    \hline
    \vspace{1mm}\\[-2mm]
    \multirow{5}*{One-stage Methods} & SMC \cite{yuan2017temporal} & 36.5 & 27.8 & 17.8 & - & -\\
    \vspace{1mm}\\[-2mm]
    & SSAD \cite{lin2017single} & 43.0 & 35.0 & 24.6 & - & -\\
    \vspace{1mm}\\[-2mm]
    & SS-TAD \cite{buch2017end} & 45.7 & - & 29.2 & - & -\\
    \vspace{1mm}\\[-2mm]
    & S3D \cite{zhang2018s3d} & 47.9 & 41.2 & 32.6 & 23.3 & 14.3\\
    \vspace{1mm}\\[-2mm]
    & GTAN \cite{long2019gaussian} & \bf{57.8} & 47.2 & 38.8 & - & -\\
    \hline
    \vspace{1mm}\\[-2mm]
    & Ours & 56.4 & \bf{50.6} & \bf{42.0} & \bf{31.2} & 19.6\\[-2mm]\\
    \toprule
  \end{tabu}
\end{table*}

\subsection{Comparison with State-of-the-art.} 
\begin{figure*}[t]
\centering
\subfigure[]{
\begin{minipage}{0.7\textwidth}
\centering
\includegraphics[width=1\textwidth]{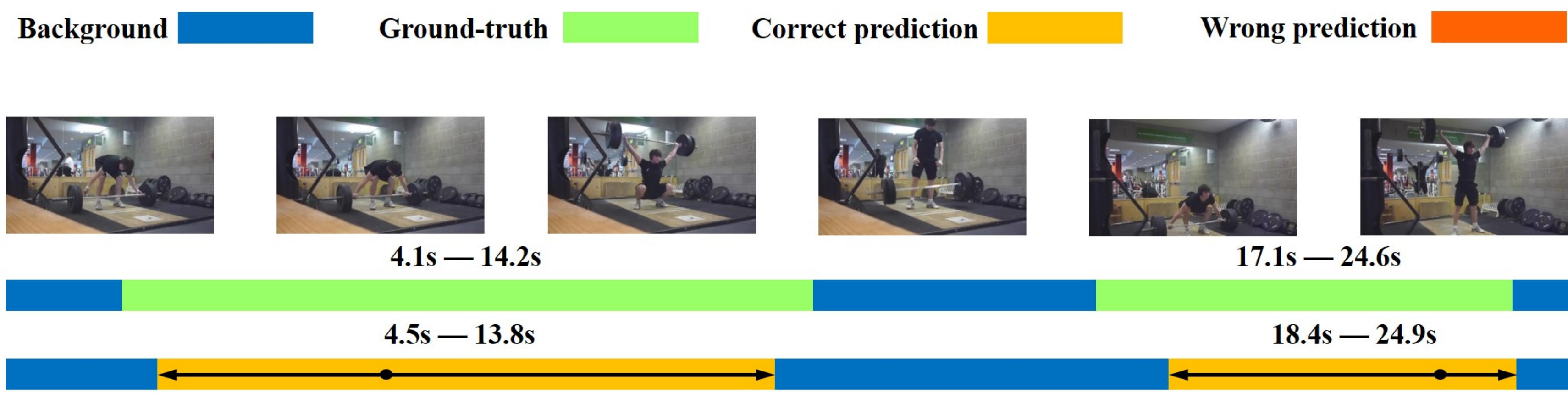}
\vspace{-2mm}
\end{minipage}
}
\vspace{1mm}\\[6mm]
\subfigure[]{
\begin{minipage}{0.8\textwidth}
\centering
\includegraphics[width=1\textwidth]{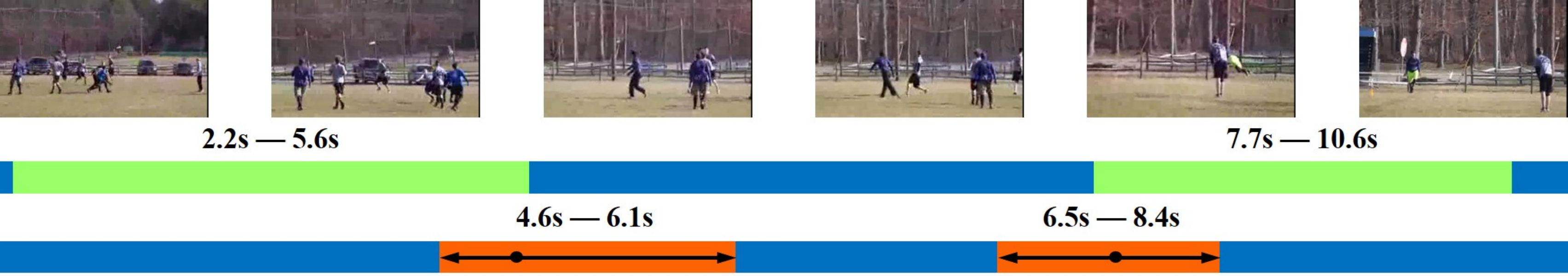}
\vspace{-2mm}
\end{minipage}
}
\vspace{1mm}\\[6mm]
\subfigure[]{
\begin{minipage}{0.8\textwidth}
\centering
\includegraphics[width=1\textwidth]{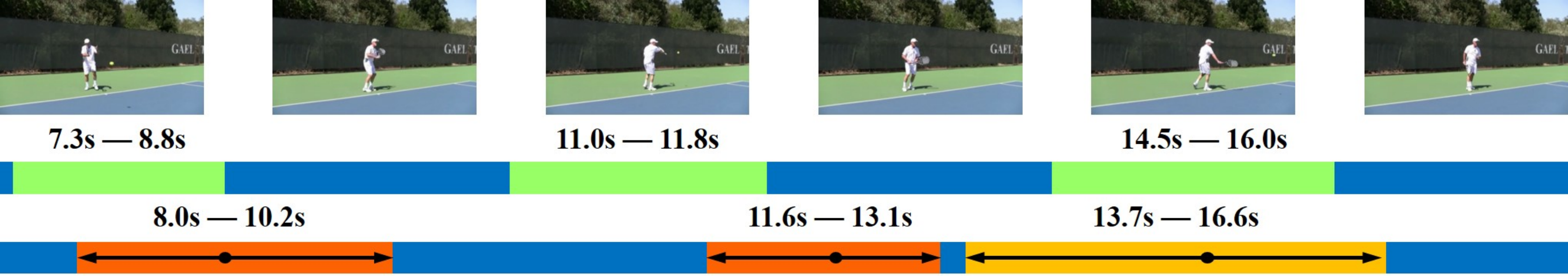}
\vspace{-2mm}
\end{minipage}
}

\caption{Visualization of predictions by AFO-TAD on three different action categories in THUMOS'14. Background are marked in blue, ground-truth instances are marked in green. Yellow denotes the correct predictions while orange for wrong predictions.}
\label{fig4}
\end{figure*}

We summarize the comparison results between AFO-TAD and other state-of-the-art methods in Table \ref{tb1} with the IoU thresholds varied from 0.3 to 0.7. The results show that with the simple C3D network, our model improves the performance to a large margin. AFO-TAD significantly outperforms other one-stage methods which also employ C3D as their backbones, like SSAD and SS-TAD. In particular, GTAN achieves 38.8\% mAP@0.5 with P3D as its backbone, while its mAP@0.5 is 37.9\% with C3D. Compare to the two-stage methods, our method outperforms BSN and BMN by increasing mAP@0.5 by 5.1\% and 3.2\% respectively. These results demonstrate the superiority of our network. Figure \ref{fig4} showcases the predictions of three video by AFO-TAD.

\subsection{Ablation Study.} In order to understand the effect of receptive field better, we evaluate AFO-TAD with different variants on THUMOS'14. In this section, we investigate how the dilated convolutional layer and deformable convolutional layer influence the detection performance.

The results of AFO-TAD using dilated convolutional layers with different dilation rates are shown in Table \ref{tb2}. Note that the convolutional layer with dilated rate 1 retrogresses to the normal convolutional layer. We can notice that the detection results are heavily influenced by dilation rate, which indicates that the performance of anchor-free one-stage method is sensitive to the receptive field. Although the performance improves with the increase of receptive field, larger receptive field doesn't always guarantee the best result. We argue that increasing the receptive field with dilated convolutional layers has several drawbacks: (1) it is hard to determine the best dilation rate, and (2) it lacks robustness with fixed dilation rate (hence the fixed receptive field) for action detection, especially for actions with large length variance.

In order to address these issues, we propose to use temporal deformable convolutional layer instead of finding the best dilation settings. Table \ref{tb3} empirically compares the performance of AFO-TAD with different number of deformable convolutional layers. It is worth noting that the performance decreases with the number of deformable layers. However, in \cite{dai2017deformable}, three deformable convolutional layers are embedded to get the best performance for object detection task. We tentatively interpret this to the fact that two-dimensional deformable convolution pays more attention on the shape or the outline of objects and it is hard to capture the shape variation with only one deformable layer. Whereas for temporal action detection, temporal deformable convolution only needs to adjust the temporal receptive field, hence requires less variations (i.e., less deformable layers).

Table \ref{tb4} showcases the importance of receptive field for one-stage action detectors. The down-sampler (max-pooling layer) provides more comprehensive feature map with larger receptive field, which increases the mAP@0.5 from 28.5\% to 39.4\%. Without any bells and whistles, the mAP@0.5 of AFO-TAD achieves the comparable performance (39.4\%) compares with other state-of-the-art methods (Table \ref{tb1}), which can be attributed to the extraordinary superiority of anchor-free one-stage methods for temporal action detection task. Although dilated convolution is beneficial to action detection in our work, the optimal dilated rate is difficult to determine. The temporal deformable convolution not only avoids the hyper-parameters of dilated convolution, but also makes prediction with dynamic receptive field instead of fixed temporal scales.

\begin{table}[!t]
  \huge
  \centering
  \scriptsize
  \caption{Results (mAP@0.5) of using the dilated convolution in the last 1 and 2 layers in RFAM.}
  \label{tb2}
  \begin{tabu}to 0.7\columnwidth{*7{X[-1, c]}}
    \\[-2mm]
    \toprule\\[-2.5mm]
    Dilated Rate & 1 & 2 & 3 & 4 \\[0.5mm]
    \hline
    \vspace{1mm}\\[-2mm]
    Last 1 layer & 39.4 & 39.6 & 41.2 & 40.1 \\
    \vspace{1mm}\\[-2mm]
    Last 2 layers & 39.4 & \bf{41.4} & 38.9 & 40.9 \\
    \vspace{1mm}\\[-2mm]
    \toprule
  \end{tabu}
\end{table}

\begin{table}[!t]
  \huge
  \centering
  \scriptsize
  \caption{Results (mAP@0.5) of using deformable convolution layers in the last 1, 2 and 3 layers in RFAM.}
  \label{tb3}
  \begin{tabu}to 0.7\columnwidth{*7{X[-1, c]}}
    \\[-2mm]
    \toprule\\[-2.5mm]
    Number of deformable convolutional layers & 1 & 2 & 3 \\[0.5mm]
    \hline
    \vspace{1mm}\\[-2mm]
    mAP@0.5(\%) & \bf{42.0} & 38.9 & 11.7 \\
    \vspace{1mm}\\[-2mm]
    \toprule
  \end{tabu}
\end{table}

\begin{table}[!t]
  \huge
  \centering
  \scriptsize
  \caption{Performance of various designs of AFO-TAD. The down-sampler is the max-pooling layer in Figure \ref{fig2}.}
  \label{tb4}
  \begin{tabu}to 0.7\columnwidth{*7{X[-1, c]}}
    \\[-2mm]
    \toprule\\[-2.5mm]
    down-sampler &  & \checkmark & \checkmark & \checkmark\\[0.5mm]
    \vspace{1mm}\\[-2mm]
    dilated layer &  &  & \checkmark & &\\[0.5mm]
    \vspace{1mm}\\[-2mm]
    deformable layer &  &  &  & \checkmark &\\[0.5mm]
    \vspace{1mm}\\[-2mm]
    \hline
    mAP@0.5(\%) & 28.5 & 39.4 & 41.2 & \bf{42.0} \\
    \vspace{1mm}\\[-2mm]
    \toprule
  \end{tabu}
\end{table}

\subsection{Inference Speed.} In addition to the detection accuracy, inference speed of a model is also an important aspect for action detector. We investigate the inference speed of several state-of-the-art architectures with speed reported in Table \ref{tb5}. Compared with two-stage methods (SCNN, DAP and R-C3D), one-stage methods (S3D and ours) have the natural advantage in detection speed due to the simple architecture. SCNN detects action instances with sliding window which is a time consuming stage. As we can see, the methods treat proposal generation and proposal classification separately operate at slow inference speed. R-C3D, similar to Faster-RCNN \cite{ren2015faster}, shares the video feature map for proposal generation and classification which accelerates the inference speed a lot. S3D, a one-stage action detector inspired by SSD \cite{liu2016ssd}, directly predicts instances without the proposal generation stage. Obviously, the one-stage architectures have great advantage in inference speed. Moreover, our proposed method (AFO-TAD) achieves the fastest inference speed by eliminating the anchor mechanism used in S3D.

\begin{table}[!t]
  \huge
  \centering
  \scriptsize
  \caption{Inference speed comparison.}
  \label{tb5}
  \begin{tabu}to 0.7\columnwidth{*7{X[-1, c]}}
    \\[-2mm]
    \toprule\\[-2.5mm]
    Network & FPS & Device \\[0.5mm]
    \hline
    \vspace{1mm}\\[-2mm]
    SCNN \cite{shou2016temporal} & 60 & - \\
    \vspace{1mm}\\[-2mm]
    DAP \cite{escorcia2016daps} & 134.1 & - \\
    \vspace{1mm}\\[-2mm]
    R-C3D \cite{xu2017r} & 1030 & TITAN XP \\
    \vspace{1mm}\\[-2mm]
    S3D \cite{zhang2018s3d} & 1271 & 1080Ti \\
    \hline
    \vspace{1mm}\\[-2mm]
    Ours (AFO-TAD) & \bf{1462} & TITAN XP \\
    \vspace{1mm}\\[-2mm]
    \toprule
  \end{tabu}
\end{table}

\section{Conclusions}
In this paper, we propose AFO-TAD architecture for temporal action detection. AFO-TAD eliminates the anchor mechanism, which is widely used in other state-of-the-art methods, by directly predicting the categories and boundaries at every temporal locations of the input video. We argue that the most important factor to one-stage methods is the receptive field. In order to verify that, we investigate multiple variants of AFO-TAD and the results show that the performance can be largely improved with the change of receptive field. However, the challenge is that a suitable receptive field for one-stage detector is hard to determine. We propose to adopt the temporal deformable convolution to adaptively adjust the receptive field instead of using fixed receptive field like previous works. This architecture predicts with dynamic receptive field, thus improves the state-of-the-art performance to a large margin on THUMOS'14.

\section*{Acknowledgement}
The research was supported by the National Natural Science Foundation of China (61571353) and the Science and Technology Projects of Xi’an, China (201809170CX11JC12).

\bibliography{mybib}
\end{document}